# MPRAD: A Multiparametric Radiomics Framework

Vishwa S. Parekh[1,3], Michael A. Jacobs*[1,2]

[1]The Russell H. Morgan Department of Radiology and Radiological Science and [2]Sidney Kimmel Comprehensive Cancer Center, The Johns Hopkins School of Medicine, Baltimore, MD 21205, USA
[3]Department of Computer Science, The Johns Hopkins University, Baltimore, MD 21208, USA

Address Correspondence To:

Michael A. Jacobs, Ph.D

The Russell H. Morgan Department of Radiology and Radiological Science and Oncology,

The Johns Hopkins University School of Medicine,

Traylor Blg, Rm 309

712 Rutland Ave, Baltimore, MD 21205

Tel:410-955-7492

Fax:410-614-1948

email: mikej@mri.jhu.edu

**Preprint**

**Abstract**

Multiparametric radiological imaging is vital for detection, characterization and diagnosis of many different diseases. The use of radiomics for quantitative extraction of textural features from radiological imaging is increasing moving towards clinical decision support. However, current methods in radiomics are limited to using single images for the extraction of these textural features and may limit the applicable scope of radiomics in different clinical settings. Thus, in the current form, they are not capable of capturing the true underlying tissue characteristics in high dimensional multiparametric imaging space. To overcome this challenge, we have developed a multiparametric imaging radiomic framework termed MPRAD for extraction of radiomic features from high dimensional datasets. MPRAD was tested on two different organs and diseases; breast cancer and cerebrovascular accidents in brain, commonly referred to as stroke. The MPRAD framework classified malignant from benign breast lesions with excellent sensitivity and specificity of 87% and 80.5% respectively with an AUC of 0.88 providing a 9%-28% increase in AUC over single radiomic parameters. More importantly, in breast, the glandular tissue MPRAD were similar between each group with no significance differences. Similarly, the MPRAD features in brain stroke demonstrated increased performance in distinguishing the perfusion-diffusion mismatch compared to single parameter radiomics and there were no differences within the white and gray matter tissue. In conclusion, we have introduced the use of multiparametric radiomics into a clinical setting.

## 1. Background

Radiomics uses texture features to define potential quantitative metrics from radiological images [1-3]. The texture features extracted are based on several properties inherent to image data, such as, gray-level distribution [4], inter-voxel relationships [5-9] and shape [10]. The goal of radiomics is to provide a quantitative framework for a "radiological" biopsy of tissue, which could be correlated to the underlying tissue biology. Reviews of several studies that have employed radiomic analysis produced encouraging results for prognosis and diagnosis of different pathologies and imaging modalities in brain, breast, lung, and prostate [1].

However, the current radiomic methods are based on extraction of textural features from a single image or volume and do not extract the textural features from multimodal or multiparametric datasets consisting of combined imaging sequences. For example, multiparametric magnetic resonance imaging (mpMRI) such as proton density (PD), T2-weighted(T2), T1-weighted(T1), diffusion-weighted (DWI), and perfusion weighted imaging (PWI). These sequences produce different soft tissue contrasts of the tissue, where each imaging sequence provides a specific representation of the tissue based on the underlying physics and gray levels. Integrating the imaging information from different radiological modalities and parameters will provide a more complete view of the underlying tissue characteristics. Correspondingly, using texture analysis on a high dimensional multi-sequence data would provide information about the "true texture" of the tissue rather than from a single specific point of view. To that end, we developed a multiparametric radiomics imaging (MPRAD) framework for extracting radiomics information from multiparametric and multimodal imaging data.

In the multiparametric setting, we define a tissue signature (TS) that encodes the imaging characteristics of different tissue types for characterization. These different imaging parameters from the tissue signature interact with each other in a high dimensional complex space forming a complex interaction network. Probing the complex interaction network could provide information that was not possible to extract using conventional radiomic methods. The MPRAD framework analyzes both the spatial distribution of the TS, in addition to the complex interaction network within a region of interest (ROI) to compute multiparametric imaging radiomic features. In this paper, we present the theory of the MPRAD framework, develop radiomic features for multi-sequence images and implement the MPRAD framework in two different clinical applications. First, we applied the MPRAD framework to multiparametric breast MRI for classification of benign from malignant lesions and compare the MPRAD results to single image radiomics. Then, the MPRAD framework was applied to multiparametric brain MRI for classification of diffusion-perfusion mismatch in stroke patients[11-15].

## 2. Materials and Methods

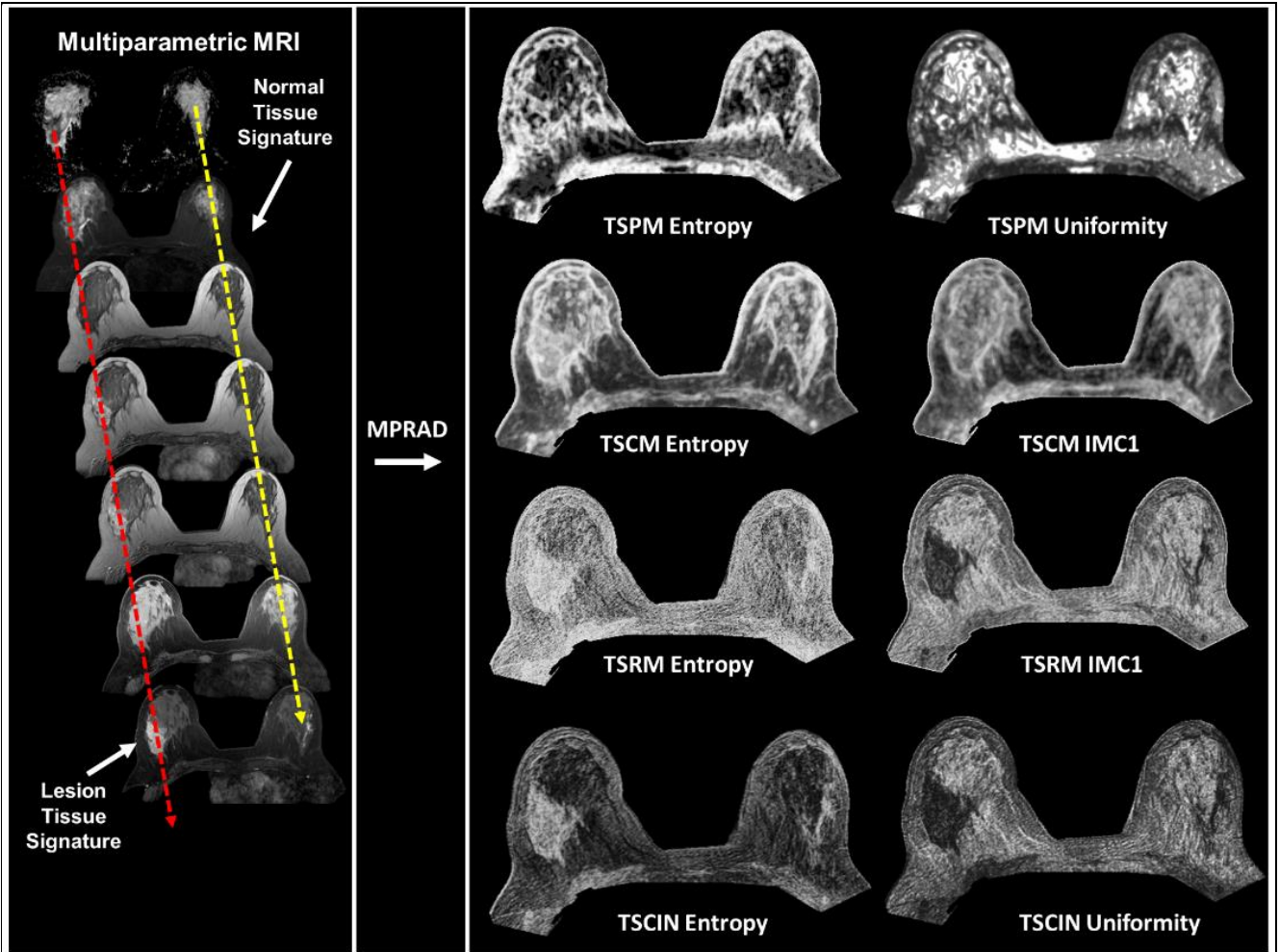

Figure 1. Illustration of the four different types of multiparametric imaging radiomic features based on first and second order statistical analysis. The tissue signature probability matrix (TSPM) and tissue signature co-occurrence matrix (TSCM) features evaluate the complex interactions between different tissue signatures while the Tissue signature complex interaction network (TSCIN) first order statistics and tissue signature relationship matrix (TSRM) features evaluate the inter-parameter complex interactions.

## Theory

### 2.1 The Tissue Signature

We define a tissue signature that represents a composite feature representation of a tissue type based on the physical modeling of the different imaging. A typical TS is shown in figure 1. Mathematically, for N different imaging parameters, a TS at a voxel position, $S_p$ is defined as a vector of gray level intensity values at that voxel position, $p$ across all the images in the data sequence for different tissue types and is given by the following equation,

$$S_p = \left[ I_p^{(1)}, I_p^{(2)}, I_p^{(3)}, \ldots, I_p^{(N)} \right]^T \quad (1)$$

Where, $I_p$ is the intensity at voxel position, p on each image.

## 2.2 tissue signature probability matrix features

The tissue signature probability matrix (TSPM) characterizes the spatial distribution of tissue signatures within a ROI. The mathematical formulation of TSPM is as follows: Suppose that the intensity values representing each voxel are quantized to $G$ levels, then the total number of possible tissue signatures in a dataset consisting of $N$ images will be equal to $G^N$. We define a function $f: T \rightarrow M$, where T is the set of all tissue signatures in the dataset and M is a N dimensional matrix with edges of length G where each tissue signature is represented as a cell. The function $f$ populates each cell of the matrix M with the frequency of occurrence of the corresponding tissue signature in the set T. The resulting matrix $M$ is called the tissue signature probability matrix (TSPM). The information content of the N dimensional multiparametric imaging dataset $(X_1, X_2, \ldots X_N)$ can be analyzed by computing the joint entropy, uniformity, and mutual information of the resultant TSPM [16]. These are defined below.

1. The TSPM entropy, $H$ is given by the following equation

$$H(X_1, X_2, \ldots X_N) = -\sum_{i_1=1}^{N_g} \sum_{i_2=1}^{N_g} \ldots \sum_{i_N=1}^{N_g} TSPM(i_1, i_2, \ldots, i_N) \log_2 TSPM(i_1, i_2, \ldots, i_N) \text{ (2)}$$

2. The TSPM uniformity, $U$ is given by the following equation

$$U(X_1, X_2, \ldots X_N) = \sum_{i_1=1}^{N_g} \sum_{i_2=1}^{N_g} \ldots \sum_{i_N=1}^{N_g} TSPM(i_1, i_2, \ldots, i_N)^2 \text{ (3)}$$

3. The TSPM mutual information, $MI$ is given by

$$MI(X_1; X_2; \ldots; X_N) = \left(H(X_1) + H(X_2) + \cdots + H(X_N)\right) - \cdots + \cdots (-1)^{N-1} H(X_1, X_2, \ldots, X_N) \text{ (4)}$$

By choosing different possible subsets $Y \subseteq \{X_1, X_2, \ldots, X_N\}$ and different values of H(Y), U(Y) and MI(Y) can be obtained producing a large number of multiparametric imaging radiomic features.

## 2.3 Tissue signature co-occurrence matrix features

The tissue signature co-occurrence matrix (TSCM) characterizes the spatial relationship between tissue signatures within a ROI. The TSCM is defined similar to the gray level co-occurrence matrix (GLCM) using two input parameters, distance (d) and angle ($\theta$) between the two tissue signature locations[6]. Mathematically, the GLCM between any two tissue signatures, $S_i$ and $S_j$ is given by the following equation

$$GLCM_d^\theta(S_i, S_j, m, n) = \left|\{r : S_i(r) = m, S_j(r) = n\}\right| \forall m, n \in \{1, 2, 3, \ldots, G\} \text{ (5)}$$

where $r \in N$ ($number\ of\ imaging\ sequences$) and $|\cdot|$ denotes the cardinality of a set.

Given a distance, $d$ and angle, $(\theta)$, the co-occurrence matrix for all such possible pairs of tissue signatures is given as follows:

$$TSCM_d^\theta(m, n) = \Sigma_{i,j} GLCM_d^\theta(S_i, S_j, m, n) \text{ (6)}$$

$$\forall\, i,j\ satisfied\ by\ d\ and\ \theta$$

Here, $TSCM_d^\theta$ is the tissue signature co-occurrence matrix. The TSCM can then be analyzed to extract twenty-two different TSCM features using the equations developed by Haralick et al [5].

## 2.4 Tissue signature complex-interaction network analysis features

The tissue signature complex interaction network (TSCIN) analysis characterizes the complex interactions that define the inter-parametric relationships between different imaging parameters using statistical analysis. The TSCIN features are extracted by transforming a high dimensional multiparametric radiological imaging data into a radiomic feature map using first or higher order statistical analysis of the tissue signature vectors, $S_p$ at each voxel position. The TSCIN feature maps are transformed into a single radiomic quantitative value corresponding to a region of interest using a summary statistical metrics mean, median or standard deviation.

### 2.4.1 First order TSCIN features

The first order TSCIN features are straightforward and calculated directly from the tissue signature. For example, the TSCIN entropy at a voxel position, p is given by the following equation:

$$Entropy_{TSCIN} = entropy\,(S_p)\ (7)$$

Similarly, all the other first order TSCIN features follow.

### 2.4.1 Second order TSCIN features

The second order TSCIN features characterize the inter-parameter relationship within the tissue signature by computing a TSCIN relationship matrix (TSRM) and are derived below. Mathematically, TSRM for a $N$ dimensional tissue signature at voxel position, $p$ with N imaging sequences quantized to $G$ gray levels is given by the following equation:

$$TSRM_d^p(i,j) = \left|\left\{k: I_p^{(k)} = i, I_p^{(k+d)} = j\right\}\right|\ \forall\, i,j \in \{1,2,3,\dots,G\}, k \in \{1,2,\dots,N\}\ (8)$$

Here, $d$ represents the distance between the two imaging parameters, $I^{(k)}$ and $I^{(k+d)}$.

The TSRM is dependent on the relative location of different imaging parameters within the tissue signature. The structure of the TSRM is similar to a G x G gray level co-occurrence matrix, thereby, allowing us to utilize all the twenty-two equations established to extract relevant features from such matrices [5]. All the four classes of the MPRAD features developed in this manuscript are illustrated in **figure 1** on an example multiparametric breast MRI dataset.

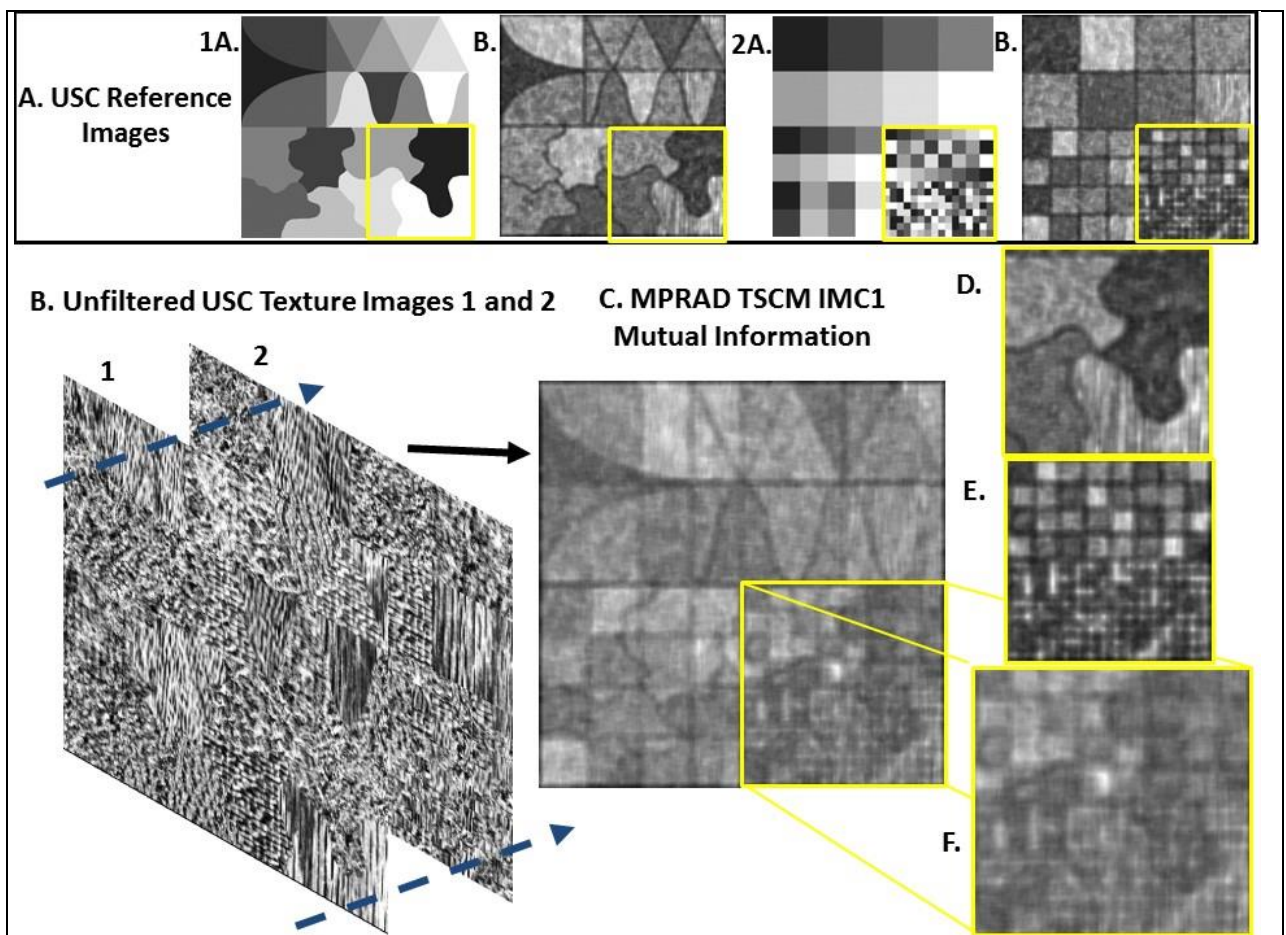

Figure 2. **A.** USC reference texture ground truth images. 1A. Reference image made out of a composite of several different shapes and textures and (1B. single radiomic image. 2A. Composite Reference image and (2B) single radiomic image **B.** Multiparametric USC composite images. **C.** mpRadiomics image of USC images. **D and E.** Enlarged radiomic images from reference images 1 and 2. **F.** Enlarged mpRadiomic image from the combination of the images. The multiparametric radiomic features were able to capture the differences in both shape and intensity distribution of both single parameter radiomic images with excellent detail of the underlying structure.

## 3. Digital and Clinical Data

### 3.1 Digital Phantom

The multiparametric imaging radiomic feature extraction methods developed in this manuscript were tested on the well-known texture phantom from University Southern California (USC) shown in **figure 2A and B** [7,17]. The texture phantom images consist of a composite mixture of several raw texture images of grass, sand, wool, water, and others derived by Brodatz [18]. Using these composite images, ground truth texture images (**figure 2-1A and 2A**) can be determined to demonstrate the effectiveness of any radiomic method [7,17]. The ground truth images are shown in the top row in **figure 2**. We stacked a series of these texture phantom images (**figure 2B1 and B2**) to demonstrate the effectiveness of the MPRAD tissue signature model to accurately segment each of the different textures. We applied single image radiomics to each image and MPRAD to stacked images to compare the results from the two methods.

## 3.2 Clinical Data

**Informed Consent:** All studies in accordance to the institutional guidelines for clinical research under IRB approved protocol by our institution for this retrospective study.

### 3.2.1 Clinical Breast Data Set: Multiparametric breast MRI dataset consisted of a cohort of 138 patients to classify between malignant and benign lesions. MRI scans were performed on a 3T magnet (Philips), using a dedicated phased array breast coil with the patient lying prone with the breast in a holder to reduce motion. Briefly, the mpMRI sequences were T1WI, T2WI, DWI, pharmacokinetics (PK) DCE (15 second resolution), and post contrast high resolution images. For registration, the DCE post contrast images were used as the reference volume. The acquisition and post processing methods for breast MRI, registration, segmentation and classification have all been detailed in [19,20,21, Parekh, 2017 #37].

### 3.2.2 Clinical brain stroke data

Our stroke data set consisted of ten patients (n=10, five women and five men, age=64±19 years) that were imaged at the acute time point (<12h) after stroke on a 1.5T clinical MRI system using a phased array head coil. The MRI parameters were: T1WI sagittal MPRAGE image (TR/TE 200/2.46ms, field of view (FOV)=24 cm x 24 cm, slice thickness (ST) = 5 mm), axial T2WI FLAIR (TR/TE/TI = 9000/105/2500ms, FOV=17.3 cm x 23 cm, ST=4 mm), axial DWI (TR/TE=9000/98ms, b-values=0 and 1000 s/mm$^2$, FOV 23 cm x 23cm, matrix=128x128, ST=4 mm) and echo planar T1WI perfusion (TR/TE = 1350/30ms, FOV=23 cm x 23cm, ST = 4 mm, total duration = 90 seconds). The contrast agent GdDTPA (Magnevist) was power injected at a dose of 0.1 mmol/kg and at a rate of 5 cc/sec. For registration, the PWI images were used as the reference volume for the DWI to be registered to.

### Area and Quantitative MRI measurements

The infarcted tissue was segmented from the PWI, DWI, and ADC map using the Eigen filter algorithm [22,23]. The input to Eigen filter algorithm was the selection of white matter, gray matter, cerebrospinal fluid and potential infarcted tissue and the tissue at risk using pixels from each tissue type and respective sequence. The Eigen filter uses a Gram-Schmidt orthonormalization to segment each tissue type. The areas for the tissue at risk and potential infracted tissue were calculated by counting the number of pixels within segmented regions followed by multiplication with the pixel resolution in mm$^2$. The ROIs defined by the Eigen filter were overlaid on the ADC map and TTP map to obtain quantitative measurements of the tissue.

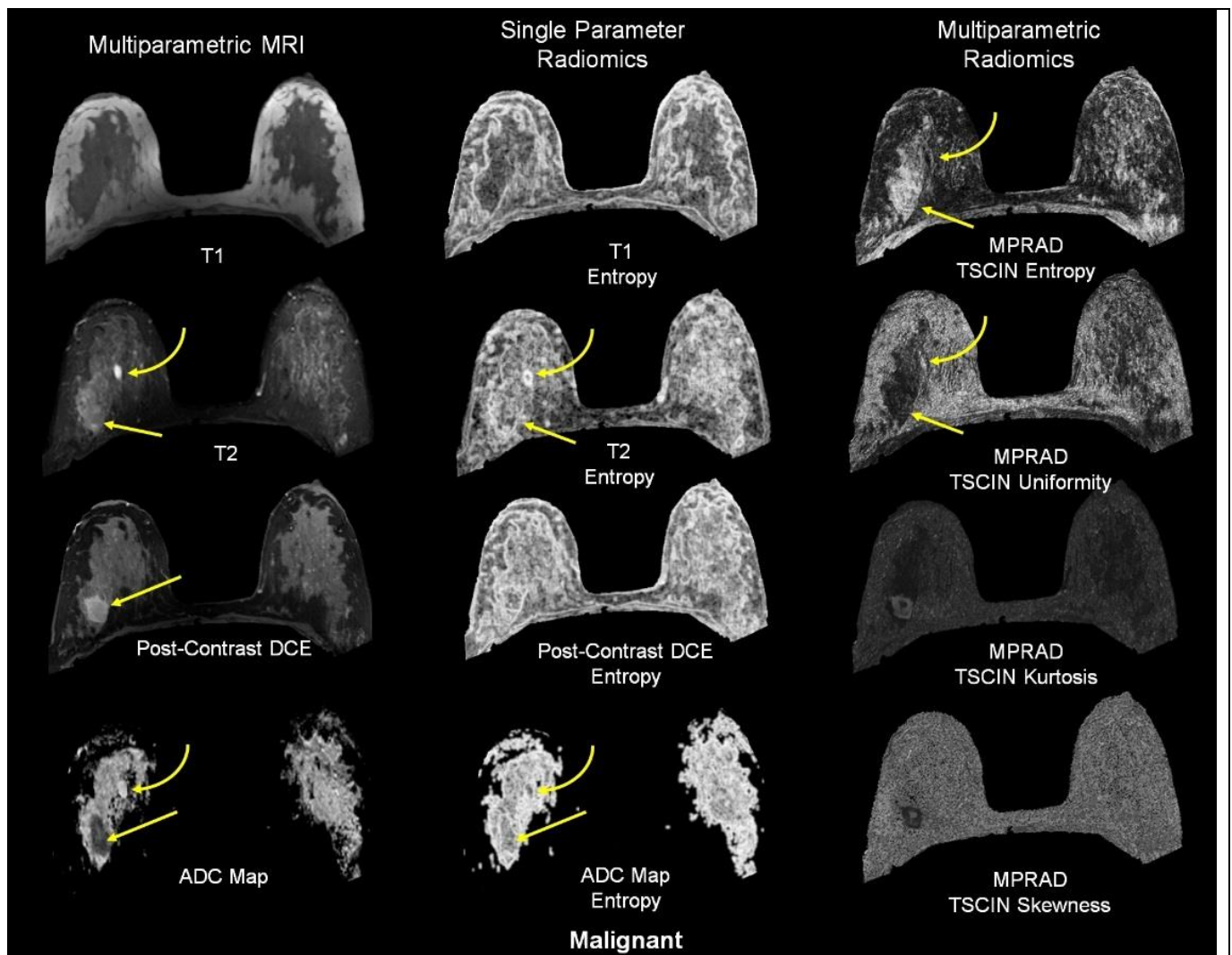

Figure 3.   The radiomic feature maps (RFM) obtained from single and multiparametric radiomic analysis in a malignant patient.  The straight yellow arrow highlights the lesion. The curved arrow demonstrates a benign cyst. **A.** Multiparametric MRI parameters **B.** Single radiomic gray level co-occurrence matrix (GLCM) entropy features maps of each MRI parameter.  **C.** The MPRAD RFMs tissue signature co-occurrence matrix (TSCM) and tissue signature complex interaction network (TSCIN) radiomic features. Note, the improved tissue delineation between the different tissue types using MPRAD

## 3.3 MPRAD Radiomic Analysis

Radiomic maps and features were computed by filtering mpMRI images with statistical kernels based on the first order TSPM and TSCIN (e.g. entropy) features and second order TSCM and TSRM features (Haralick's gray level co-occurrence matrix features) described above. The optimal neighborhood and gray level quantization values for filtering were determined by the image resolution, bit depth of the radiological images and empirical analysis of the uniformity and noise within the radiomic maps. The radiomic parameters of neighborhood and gray level quantization were set to 15x15 with 256 gray levels for synthetic USC texture images, 5x5 and 128 gray levels for breast mpMRI and 3x3 and 32 gray levels for brain mpMRI (DWI and PWI). The ROIs from the different tissue types were segmented and overlaid on the MPRAD maps for quantification of the texture values.

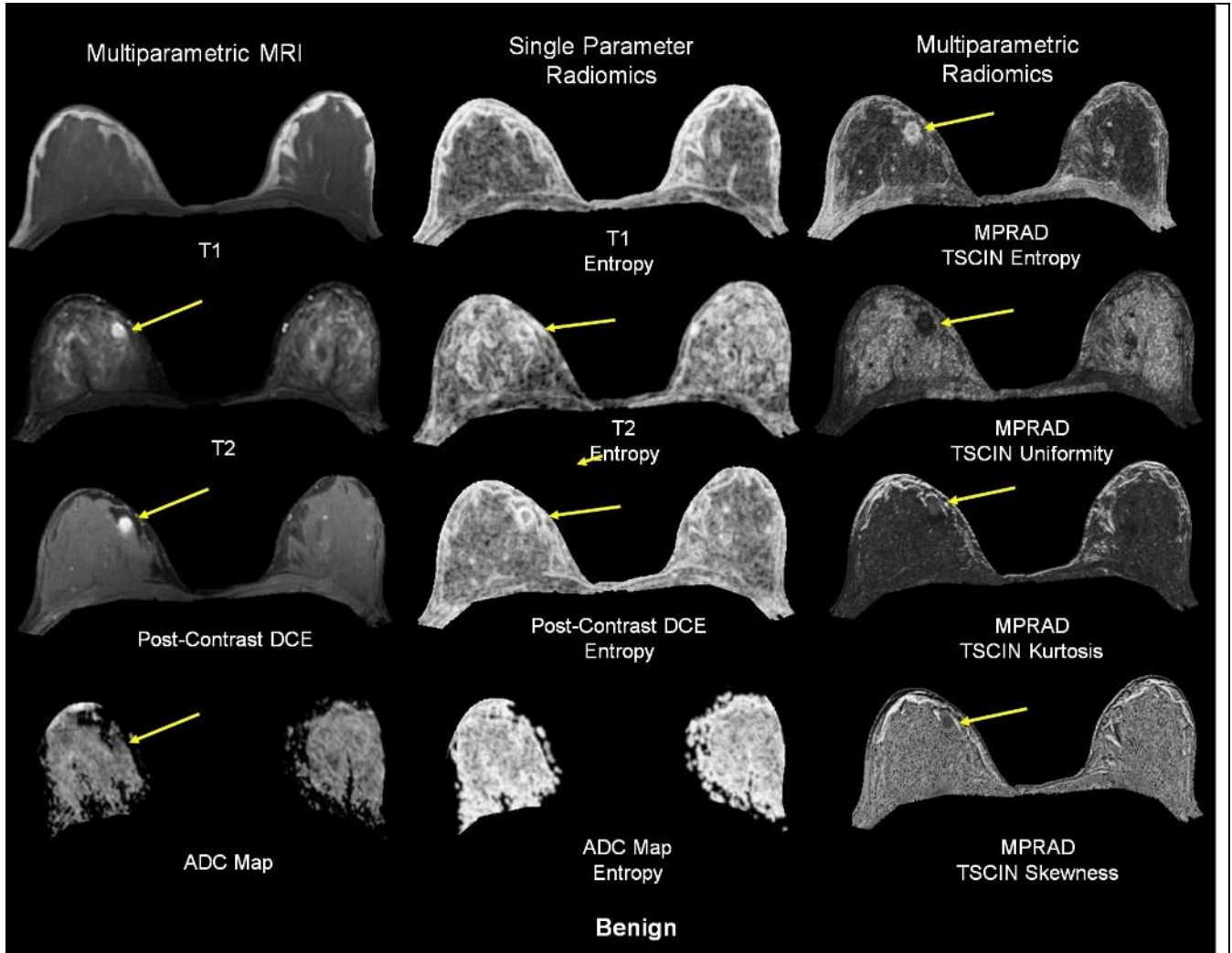

Figure 4. The radiomic feature maps (RFM) obtained from single and multiparametric radiomic analysis in a benign patient. The yellow arrow highlights the lesion. A. Multiparametric MRI parameters B. Single radiomic gray level co-occurrence matrix (GLCM) entropy features maps of each MRI parameter. C. The MPRAD RFMs tissue signature co-occurrence matrix (TSCM) and tissue signature complex interaction network (TSCIN) radiomic features.

### 3.4 IsoSVM

We developed a new feature embedding and classification framework termed IsoSVM [24] by modifying two machine learning algorithms, Isomap and support vector machine (SVM) [25,26]. This hybrid IsoSVM classifier enables accurate classification of high dimensional data sets. Briefly, the Isomap is a non-linear dimension reduction algorithm based on the geodesic distance and multidimensional scaling. The SVM is a linear binary classification algorithm that attempts to create a hyperplane that best separates the different groups. The application of Isomap algorithm prior to SVM transforms the high dimensional MPRAD feature space into a linearly separable space. Then, the SVM algorithm trains a classification model to classify between benign and malignant patients on the transformed feature space. The imbalance in the number of benign and malignant patients was resolved by setting a higher misclassification cost for benign than malignant patients.

### 3.5 Statistical analysis

Summary statistics (mean and standard deviations) were calculated for each MPRAD feature. An unpaired t-test (two-sided) was performed to compare the MPRAD features obtained for different tissue types. Univariate logistic regression analysis was used to find associations between the four MPRAD features and the final diagnosis. Sensitivity, specificity and Receiver Operating Characteristic (ROC) curve analysis was performed to assess the diagnostic performance of each parameter in the breast data set. Statistical significance was set at p<0.05.

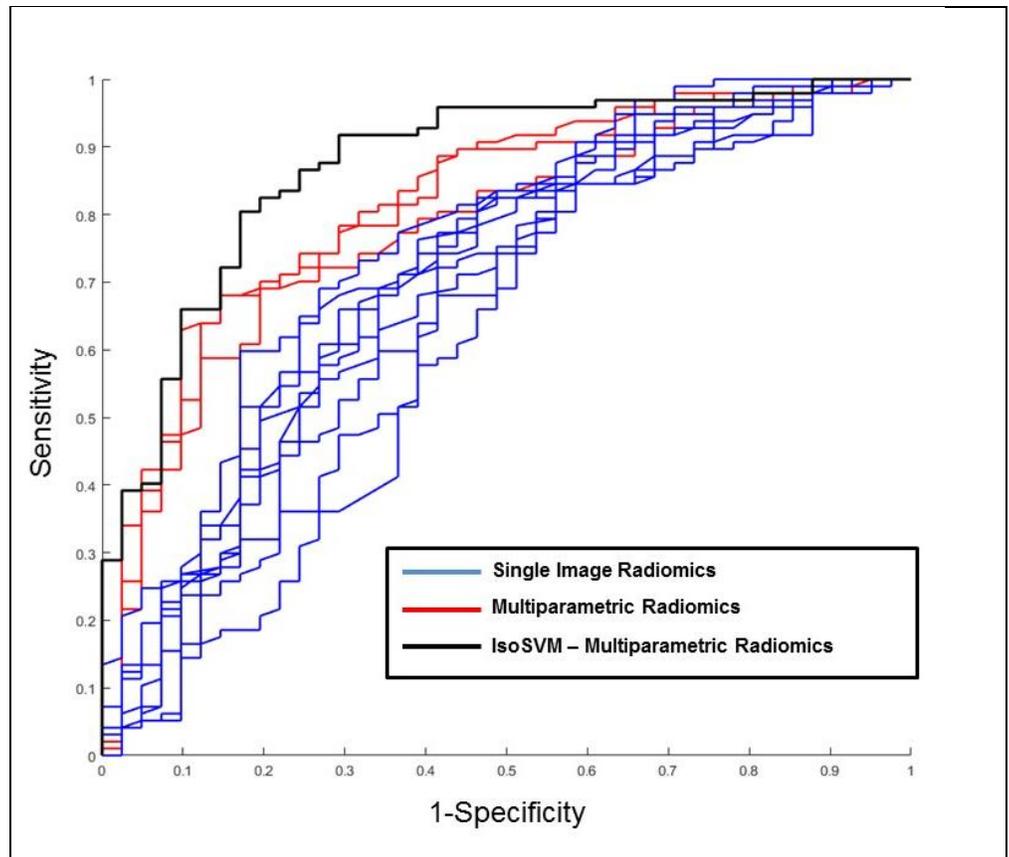

Figure 5. Comparison between the predictive accuracy of the single parameter based radiomics features and multiparametric radiomic features using receiver operating characteristic (ROC) curve analysis. The multiparametric radiomic feature ROC curves (displayed in **red**) produced area under the ROC curve (AUC) values that were 9%-28% greater than the AUCs obtained for single parameter radiomics (ROC curves displayed in **blue**). The ROC curve obtained from applying IsoSVM for classification of benign from malignant patients is displayed in **black**. The area under the ROC curve (AUC) for IsoSVM was obtained at 0.87.

## 4. Results

### 4.1 Digital Texture Phantom

The MPRAD features results on each USC composite texture image are shown in **figure 2**. The reference texture ground truth images provide a method to evaluate the radiomic features from known objects with high texture. From the two composite images, both single and multiparametric radiomic features were able to produce a 100 percent match with the reference images confirming the method on an independent data set (figure 2 top row). Moreover, the multiparametric radiomic features were able to capture the differences in both shape and intensity distribution of both single parameter radiomic images with excellent detail of the underlying structure. The higher order entropy (GLCM) values for the highlighted subregion in **figure 2** were 9.9 and 9.7 for the single radiomic images corresponding to the square and mosaic respectively while the MPRAD TSCM entropy was 10.23.

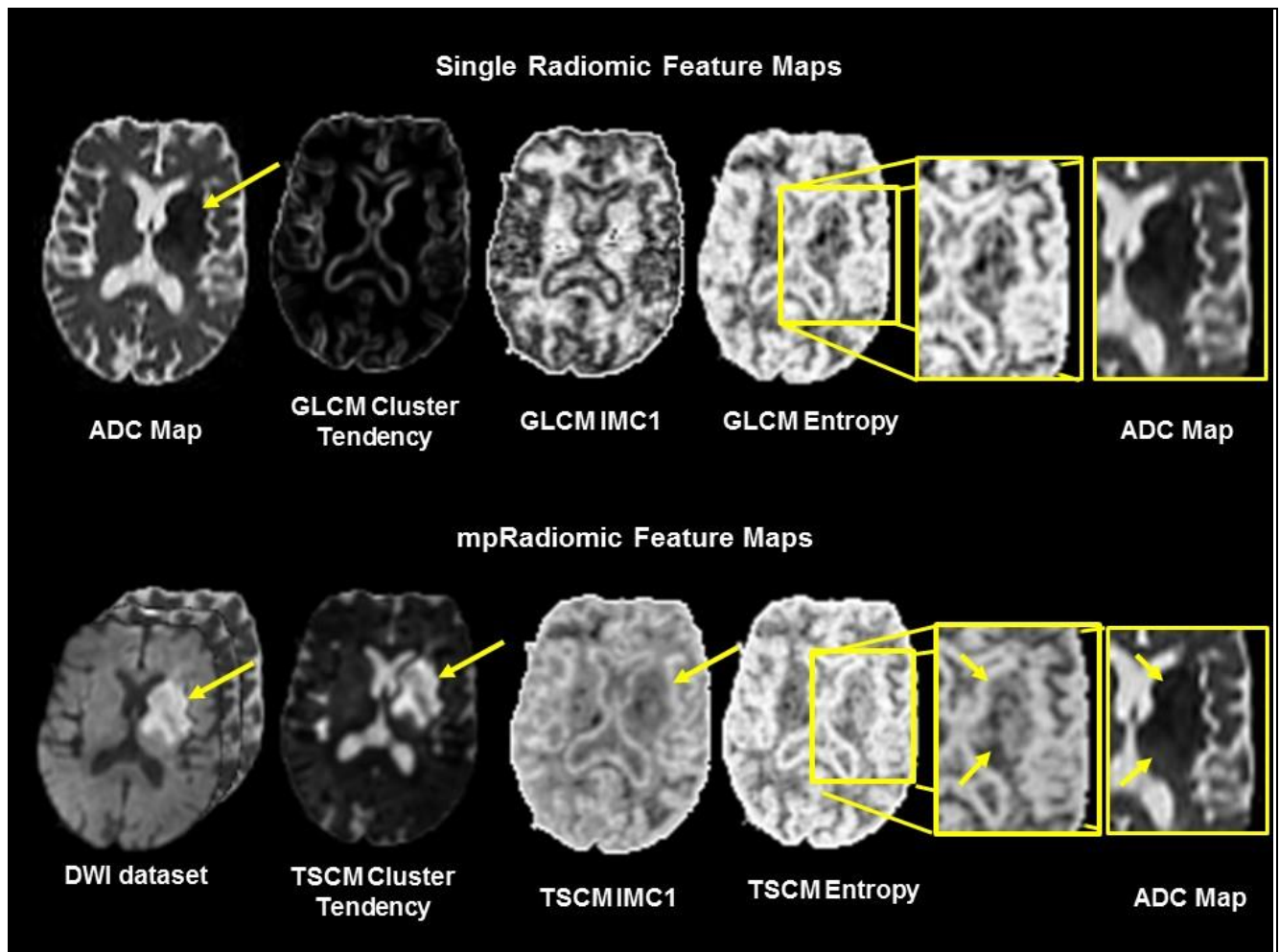

Figure 6. Illustration of radiomic feature maps (RFM) obtained from single and multiparametric radiomic analysis of an acute stroke patient with mpMRI Diffusion weighted imaging and ADC mapping. **Top Row.** ADC map with the yellow arrow showing the densely ischemic tissue. The RFMs in the illustrate different gray level co-occurrence matrix (GLCM) single radiomic features maps for the ADC map. The delineation of the infarcted tissue is hard to discern. **Bottom Row.** MPRAD of the DWI data set with yellow arrows showing the infarcted tissue. The MPRAD demonstrates excellent delineation of the infarcted tissue. The enlarged area shows the heterogeneity of the lesion.

## 4.2 Multiparametric breast MRI

The set of multiparametric imaging radiomics (MPRAD) features were extracted from one hundred and thirty-eight women with breast lesions that underwent mpMRI scans. The mean age of the patients was 52±11 ranging between 24-80. Of the 138 patients, there were 97 patients with biopsy proven malignancy whereas 41 patients had benign lesions. **Figure 3** illustrates both the single and MPRAD feature maps from a patient with a malignant lesion in the upper outer quadrant of the right breast with a benign appearing cyst superior and more medial to the lesion (curved yellow arrow). The cyst is uniformly bright on T2 and the ADC map consistent with known tissue characteristics associated with cysts. Similarly, the cyst is dark on T1 and no contrast enhancement on the DCE image indicating no vascularity. Moreover, the lesion appears to heterogenous on all the MRI images with decreased ADC and increased DCE characteristics. The single radiomic images do show some texture features, however, there is a striking difference in the textural representation shown by the MPRAD radiomics. In particular, the cyst is shown with decreased entropy in the MPRAD compared to single radiomic images. The lower entropy in the cyst is consistent with the fact, that the homogenous object has less disorder and hence lower entropy. This is clearly evident when looking at the lesion which has much higher entropy values.

**Figure 4** illustrates both the single and MPRAD feature maps from a benign patient. There was a clear difference between the textural representation of the lesion and glandular tissue. Furthermore, the tissue characterization of lesion and glandular tissue was consistent for both the benign and the malignant patient. **Table 1** summarizes the quantitative values from single parameter and TSPM entropy for individual and MPRAD features on benign and malignant patient cohorts demonstrating improved tissue characterization using MPRAD. The MPRAD TSPM entropy computed using all the MRI parameters was significantly different between benign and malignant lesions (Benign: 7.06±0.27, Malignant: 8.93±0.17, p<0.00001). Furthermore, the univariate AUC of TSPM entropy was 0.82, 9% higher than the maximum AUC (0.75 for post contrast DCE) obtained from univariate analysis of first order entropy computed from different imaging parameters. More importantly, there were no significant differences between the single and multiparametric radiomic features values in the contralateral glandular tissue from benign and malignant patients as shown in **table 2**.

The top MPRAD features for differentiating benign from malignant patients have been summarized in **table 3**. Using IsoSVM with leave-one-out cross validation, these top MPRAD features produced a sensitivity and specificity of 82.5% and 80.5% respectively with an AUC of 0.87. The optimal IsoSVM parameters were k=20, d=1 with an imbalance ratio of 3:1 of benign to malignant. The predictive power of single, MPRAD and the IsoSVM model are shown in **figure 5**. The resulting ROC curves demonstrated superior discrimination between benign and malignant patients by the MPRAD compared to single radiomic as displayed in **figure 5**.

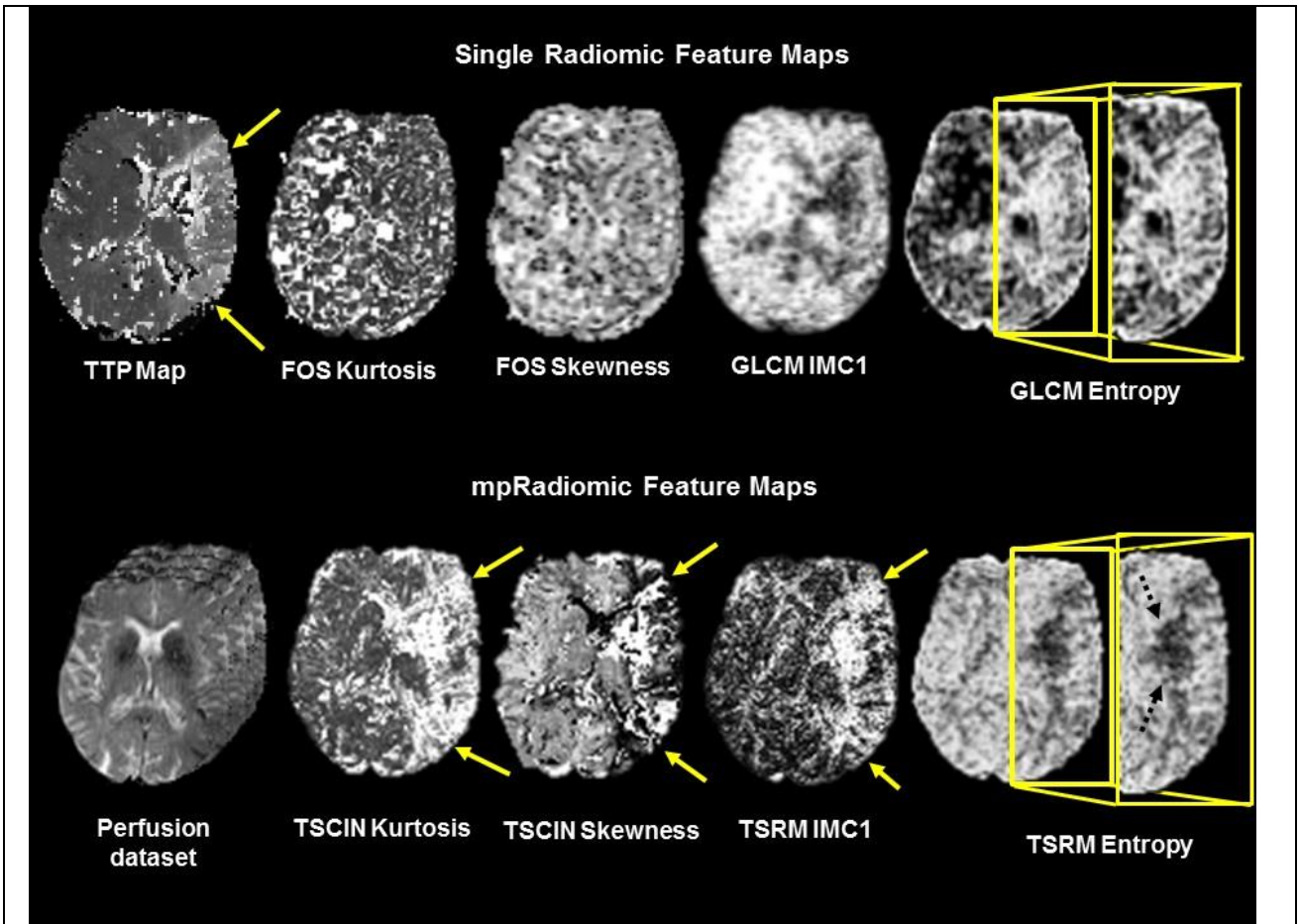

Figure 7. Illustration of radiomic feature maps (RFM) obtained from single and multiparametric radiomic analysis of an acute stroke patient with mpMRI perfusion weighted imaging. **Top Row.** Time to Peak (TTP) map from the perfusion MRI with the yellow arrow showing potential "tissue at risk". The first order (FOS) RFMs illustrate the different gray level single radiomic and co-occurrence matrix (GLCM) maps from the TTP. **Bottom Row.** MPRAD images from perfusion MRI illustrates the power mpRadiomics and the striking difference in the "tissue at risk" delineation in both the tissue signature matrix (TSCIN) and tissue signature relationship matrix (TSRM) radiomic features. The black dotted arrows show the infarcted tissue in the caudate putamen and internal capsule.

## 4.3 Multiparametric brain MRI

Four MPRAD features were extracted from ten patients with stroke imaged at acute timepoint (<12h). The mean age of patients was 64±19 ranging between 36-87. **Figures 6 and 7** illustrates both the single and multi-parametric radiomic feature maps from the DWI and PWI MRI. As shown in **figure 6**, there was a striking difference in the textural representation between the single and multi-parametric radiomics images from the DWI and ADC maps. The majority of the single radiomic second order features (GLCM) did not show any significant textural difference between infarcted tissue and tissue at risk on the ADC map. Whereas the same second order multiparametric radiomic features (TSPM) were significantly different for multiparametric complete DWI dataset. These results have been tabulated in **table 4**.

Similarly, multiparametric radiomic values for the TTP and PWI dataset (all parameters, dimensionality >50) demonstrated excellent results for the MPRAD as shown in **Figure 7**. For example, The MPRAD TSPM Entropy exhibited significant difference between infarcted tissue and potential tissue-at-risk: (6.6±0.5 vs 8.4±0.3, p=0.01). These results are summarized in **table 5** detailing the comparison between the single and multiparametric radiomics on PWI. **Table 6** summarizes the results from multiparametric radiomics applied to all the parameters.

## 5. Discussion

We have developed and validated a novel multiparametric imaging radiomics (MPRAD) framework that integrates all the data to define different tissue characteristics of the data. MPRAD outperformed single radiomic features in both synthetic and clinical datasets. The MPRAD features captured the underlying tissue texture based on tissue signatures rather than individual imaging parameter intensities. In addition, MPRAD produces full texture images for visualization of normal and lesion heterogeneity, thereby providing radiologists with a new tool for visualization and quantization of the true underlying tissue heterogeneity in conjunction with traditional images.

In multiparametric imaging settings, single radiomic features from each individual image can result in large numbers of texture features creating a high dimensional dataset across all images for analysis. These single radiomic features may not reflect the true underlying tissue contrast, heterogeneity or homogeneity and only provide a limited information corresponding to the physical modeling of each imaging parameter. The MPRAD framework extracts radiomic features that consider the complete multiparametric dataset, hence producing more meaningful features and textural visualization of the underlying tissue. In addition, the MPRAD framework allows us to analyze the complex interactions between different imaging parameters, consequently opening up a completely new source of information that did not exist with conventional radiomic features.

In breast, consistent with other reports, malignant lesions had increased entropy compared to benign lesions [1]. Importantly, there were no differences in the normal glandular tissue between each group [24]. The MPRAD radiomic features delineated different tissue types better than the single radiomic features, for example, in cysts, normal, and peri-tumoral regions. Finally, the MPRAD demonstrated excellent sensitivity and specificity with increased AUC compared to single radiomic features comparable with those achieved by radiologists.

In stroke patients, predicating stroke outcome is very challenging and the evaluation and treatment are determined by the ability to identify the ischemic penumbra, oligemic and potential salvageable tissue. Ischemic penumbra refers to brain tissue which is at risk for infarction [11,12,27,28]. Damage to this salvable tissue can be potentially prevented or reversed using thrombolytic therapy [14]. The brain tissue corresponding to ischemic penumbra and oligemia can be identified using advanced MRI parameters of diffusion-weighted imaging (DWI) and perfusion-weighted imaging (PWI) [15,28,29]. the MPRAD was able to accurately separate the diffusion-perfusion mismatch at the acute time point. The diffusion-perfusion mismatch maybe indicative of the extent of the stroke and if therapeutic intervention needed and MPRAD will provide new quantitation and visualization tools for use in stroke patients.

The integration of advanced MPRAD features with the ADC map and perfusion metrics could provide important information about the spatial distribution and characteristics of the tissue. For example, the ADC map and entropy values for infarcted tissue were decreased. The perfusion values were prolonged and where the entropy values were higher and maybe able to separate out the oligemic tissue. This is consistent with the known biology of ischemic tissue, where the tissue is dead or dying and will exhibit a more uniform pattern. In contrast, tissue at risk is highly variable and mixed with oligemic tissue with different tissue characteristics and has increased ADC and perfusion values [15]. The MPRAD radiomic entropy values were increased, reflecting this tissue heterogeneity. Thus, by combining these new radiomic features could be very helpful in clinical decision to give treatment or withhold it

In general, multiparametric imaging for applications such as brain, breast and prostate MRI produces a large number of images (>50) corresponding to each slice location, thereby producing a high dimensional image space. Extracting radiomic features from each image in such datasets may not provide complete information about the tissue. In addition, given the high dimensionality of the dataset, there is an increase in both the computational and space complexity of the radiomic process making radiomics analysis impractical for such cases. The MPRAD framework resolves this issue by extracting radiomic features that not only analyze the progression of tissue texture with time but also evaluate the overall tissue texture in large data sets.

There are, however, some technical limitations to the use the MPRAD in practice. First, there is a need for graphical processor units (GPU) with large memory and user-friendly software for processing. These items may not be currently widely available. More specific to the present study, any assessment of the clinical value of MPRAD network will require additional prospective studies. These studies would have subsequent follow-up and pathological correlation using MPRAD in breast. In stroke, these types of studies would provide us with new MPRAD data to potential predict final infraction volume or identify markers of hemorrhagic transformation overtime.

In conclusion, we have demonstrated that MPRAD framework shows a excellent potential in analysis of textural information using multimodal and multiparametric imaging in different clinical settings. With increasing use of multiparametric imaging in clinical setting, MPRAD provides an ideal framework for future clinical decision support systems.

**6. Code availability**

Our software will be freely available to academic users after issue of pending patents and a materials research agreement is obtained from the university. Due to University regulations, any patent pending software is not available until a patent is issued.

**7. Data availability**

All relevant clinical data are available upon request with adherence to HIPPA laws and the institutions IRB policies.

**8. Acknowledgments:**
National Institutes of Health (NIH) grant numbers: 5P30CA006973 (Imaging Response Assessment Team - IRAT), U01CA140204, 1R01CA190299, and The Tesla K40s used for this research was donated by the NVIDIA Corporation.

**9. Author Contributions:**
MAJ and VSP developed the concept, performed the testing, algorithm development, statistical methods, and manuscript writing.

**10. Conflict of Interest:**
The authors have no conflict of interests.

**Figures**

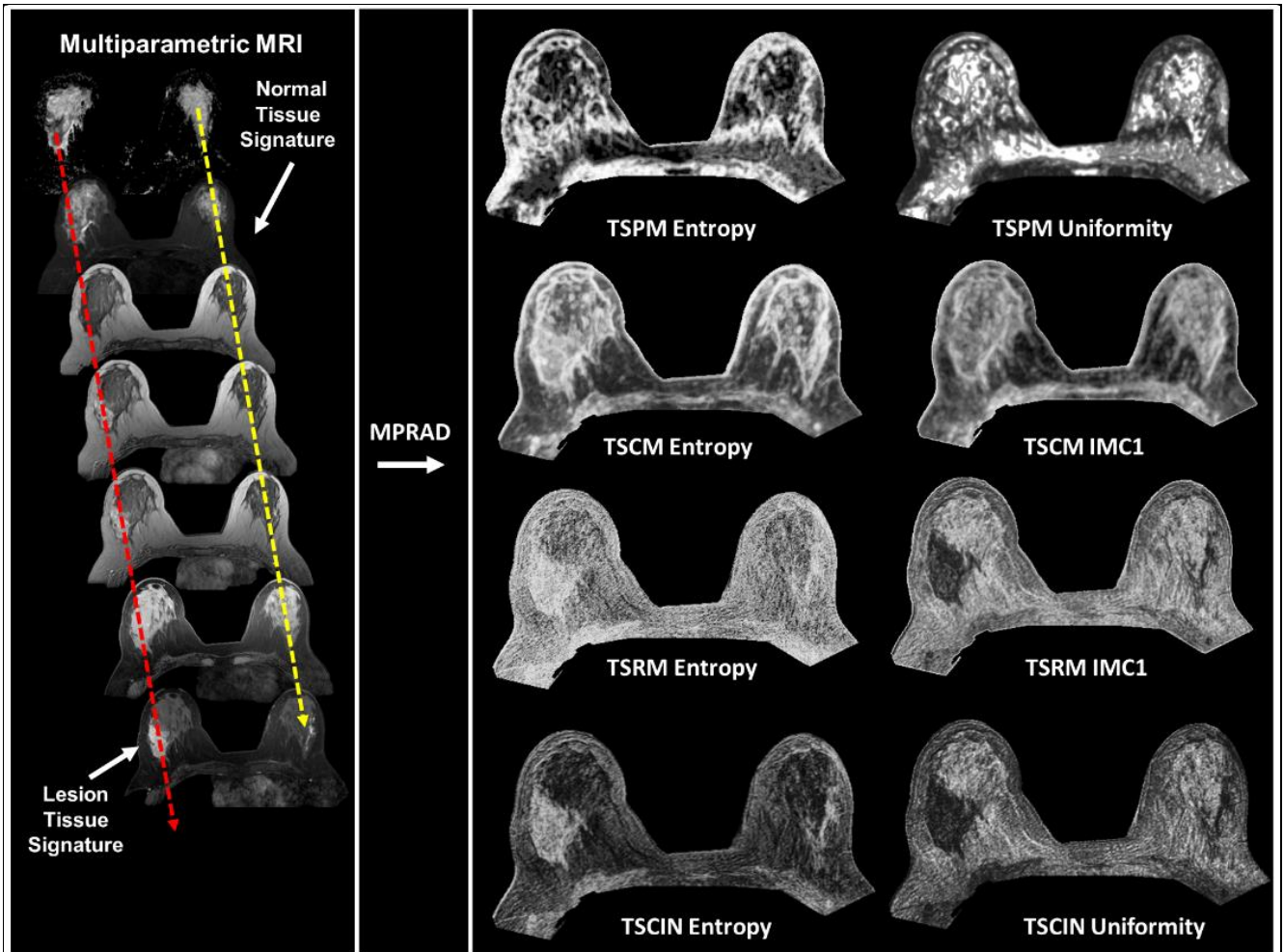

Figure 1.  Illustration of the four different types of multiparametric imaging radiomic features based on first and second order statistical analysis. The tissue signature probability matrix (TSPM) and tissue signature co-occurrence matrix (TSCM) features evaluate the complex interactions between different tissue signatures while the Tissue signature complex interaction network (TSCIN) first order statistics and tissue signature relationship matrix (TSRM) features evaluate the inter-parameter complex interactions.

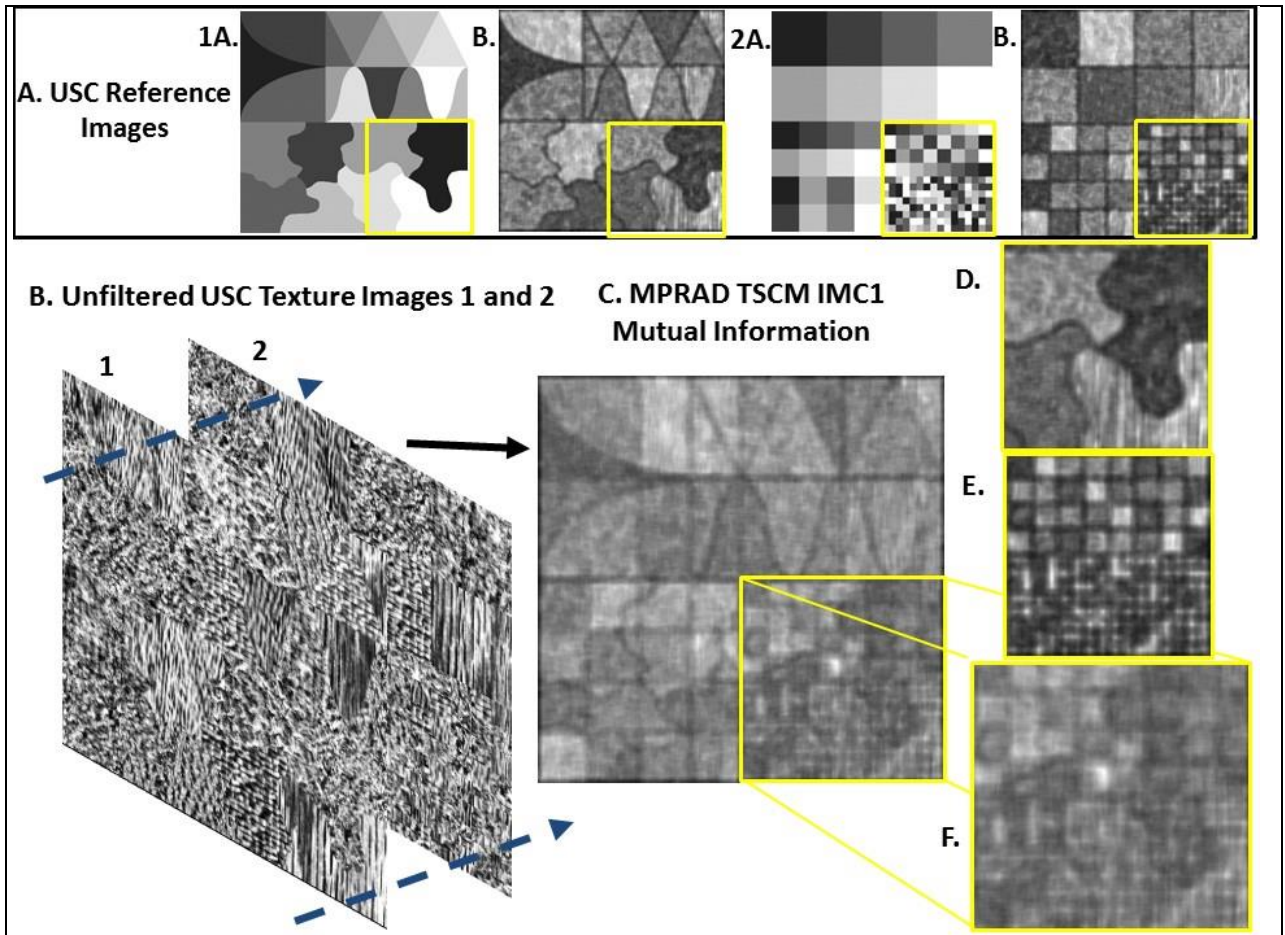

Figure 2. **A.** USC reference texture ground truth images. 1A. Reference image made out of a composite of several different shapes and textures and (1B. single radiomic image. 2A. Composite Reference image and (2B) single radiomic image **B.** Multiparametric USC composite images. **C.** mpRadiomics image of USC images. **D and E.** Enlarged radiomic images from reference images 1 and 2. **F.** Enlarged mpRadiomic image from the combination of the images. The multiparametric radiomic features were able to capture the differences in both shape and intensity distribution of both single parameter radiomic images with excellent detail of the underlying structure.

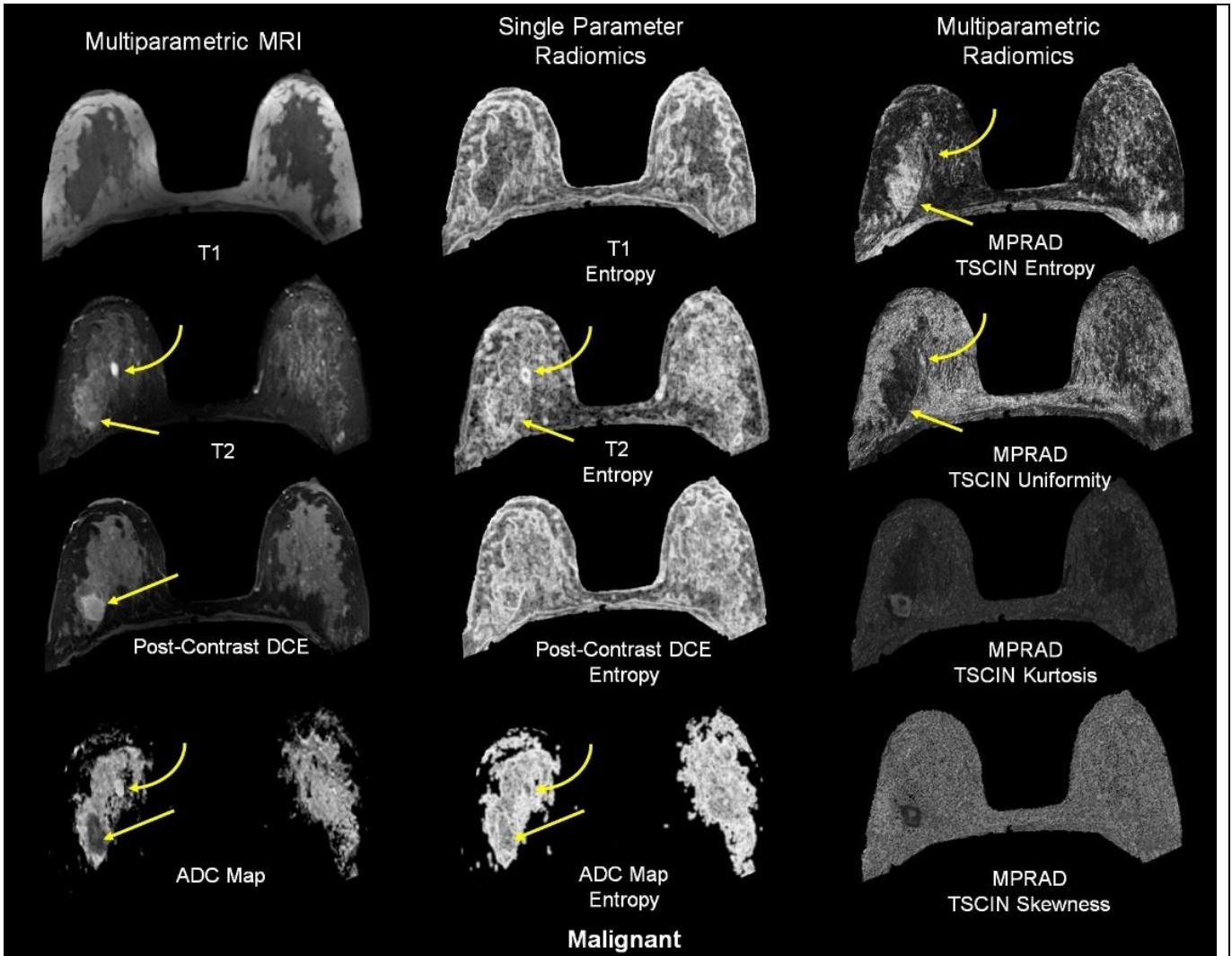

Figure 3. The radiomic feature maps (RFM) obtained from single and multiparametric radiomic analysis in a malignant patient. The straight yellow arrow highlights the lesion. The curved arrow demonstrates a benign cyst. **A.** Multiparametric MRI parameters **B.** Single radiomic gray level co-occurrence matrix (GLCM) entropy features maps of each MRI parameter. **C.** The MPRAD RFMs tissue signature co-occurrence matrix (TSCM) and tissue signature complex interaction network (TSCIN) radiomic features. Note, the improved tissue delineation between the different tissue types using MPRAD

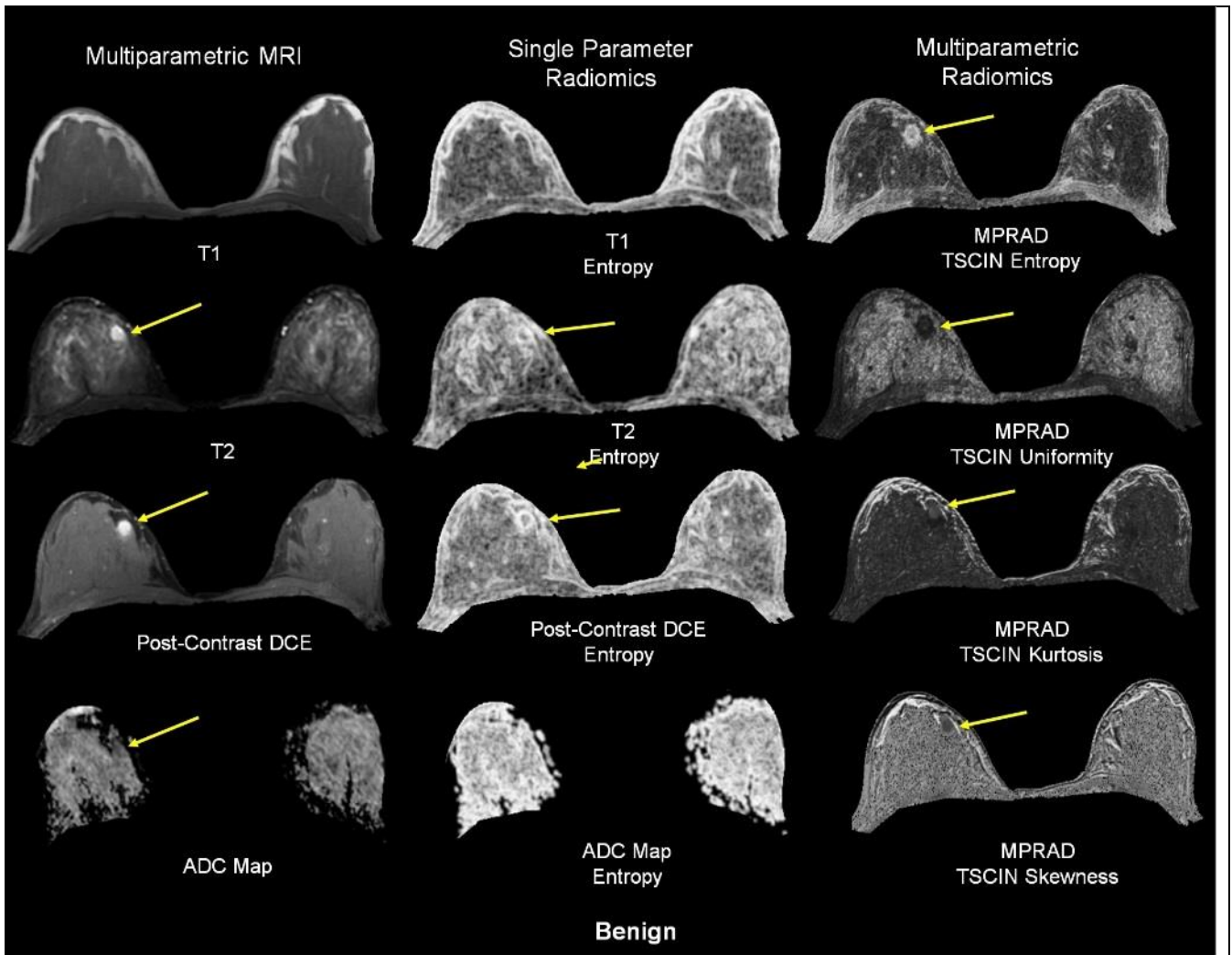

Figure 4.    The radiomic feature maps (RFM) obtained from single and multiparametric radiomic analysis in a benign patient.    The yellow arrow highlights the lesion. A. Multiparametric MRI parameters B. Single radiomic gray level co-occurrence matrix (GLCM) entropy features maps of each MRI parameter.  C. The MPRAD RFMs tissue signature co-occurrence matrix (TSCM) and tissue signature complex interaction network (TSCIN) radiomic features.

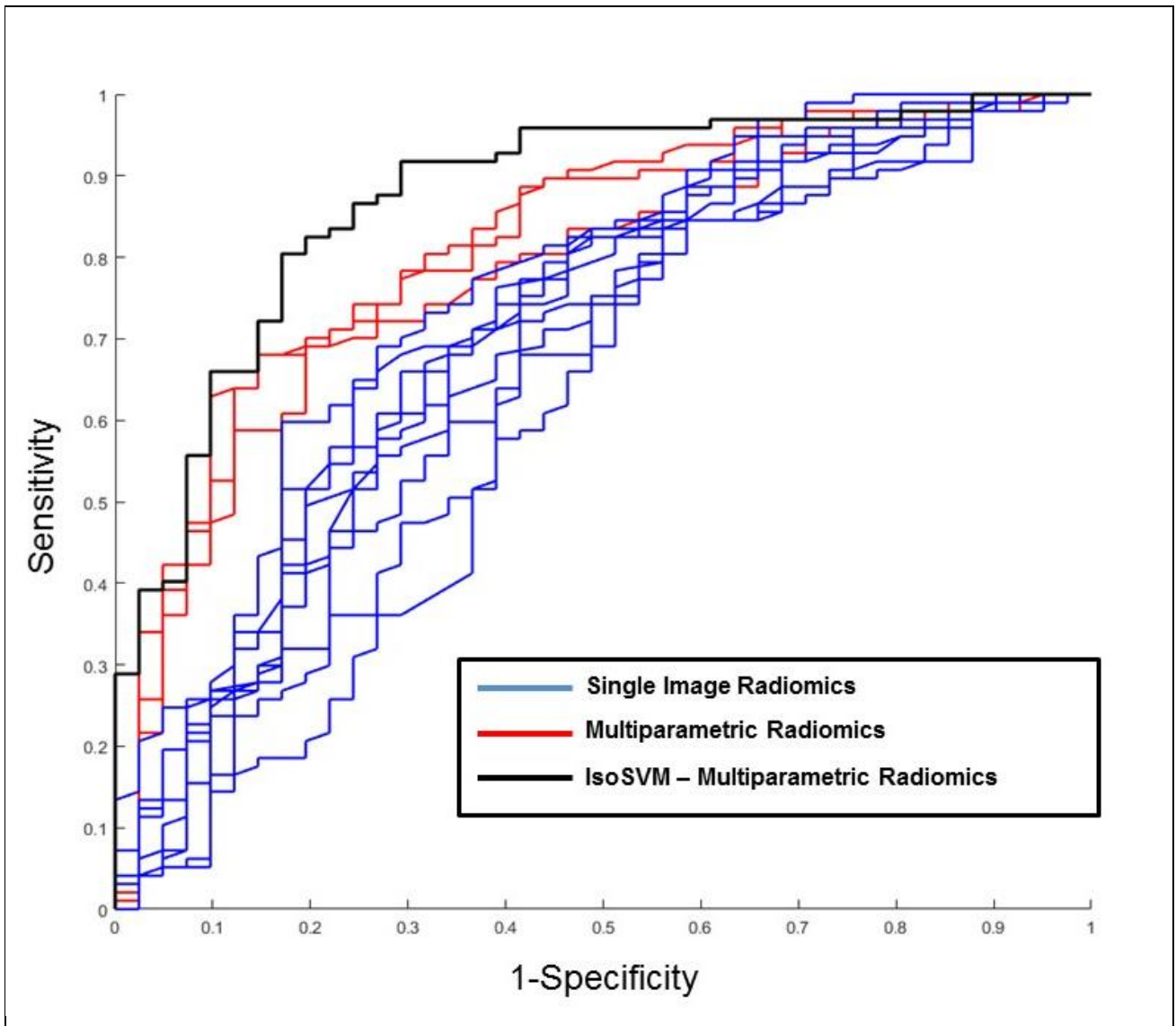

Figure 5. Comparison between the predictive accuracy of the single parameter based radiomics features and multiparametric radiomic features using receiver operating characteristic (ROC) curve analysis. The multiparametric radiomic feature ROC curves (displayed in **red**) produced area under the ROC curve (AUC) values that were 9%-28% greater than the AUCs obtained for single parameter radiomics (ROC curves displayed in **blue**). The ROC curve obtained from applying IsoSVM for classification of benign from malignant patients is displayed in **black**. The area under the ROC curve (AUC) for IsoSVM was obtained at 0.87.

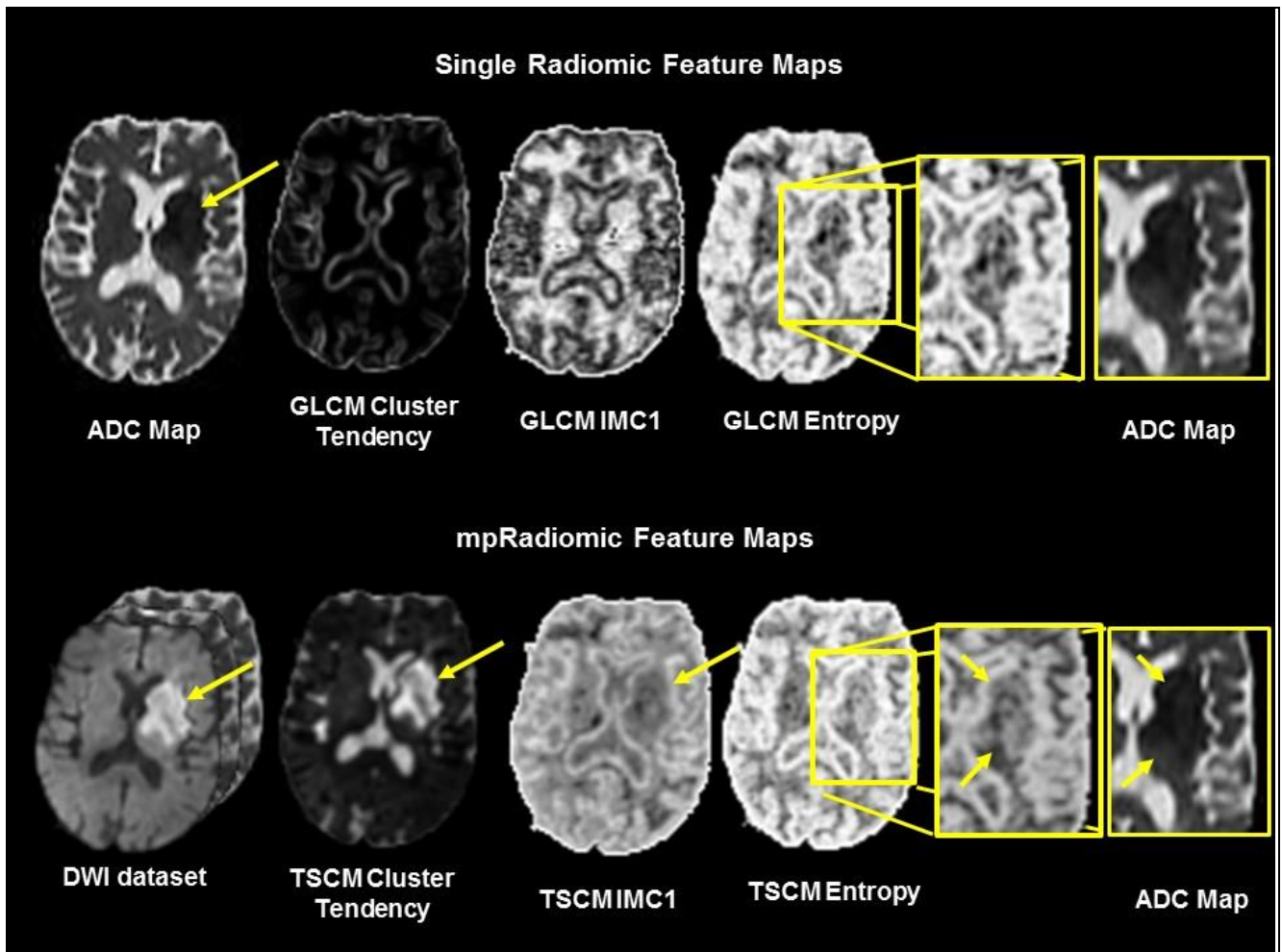

Figure 6. Illustration of radiomic feature maps (RFM) obtained from single and multiparametric radiomic analysis of an acute stroke patient with mpMRI Diffusion weighted imaging and ADC mapping. **Top Row.** ADC map with the yellow arrow showing the densely ischemic tissue. The RFMs in the illustrate different gray level co-occurrence matrix (GLCM) single radiomic features maps for the ADC map. The delineation of the infarcted tissue is hard to discern. **Bottom Row.** MPRAD of the DWI data set with yellow arrows showing the infarcted tissue. The MPRAD demonstrates excellent delineation of the infarcted tissue. The enlarged area shows the heterogeneity of the lesion.

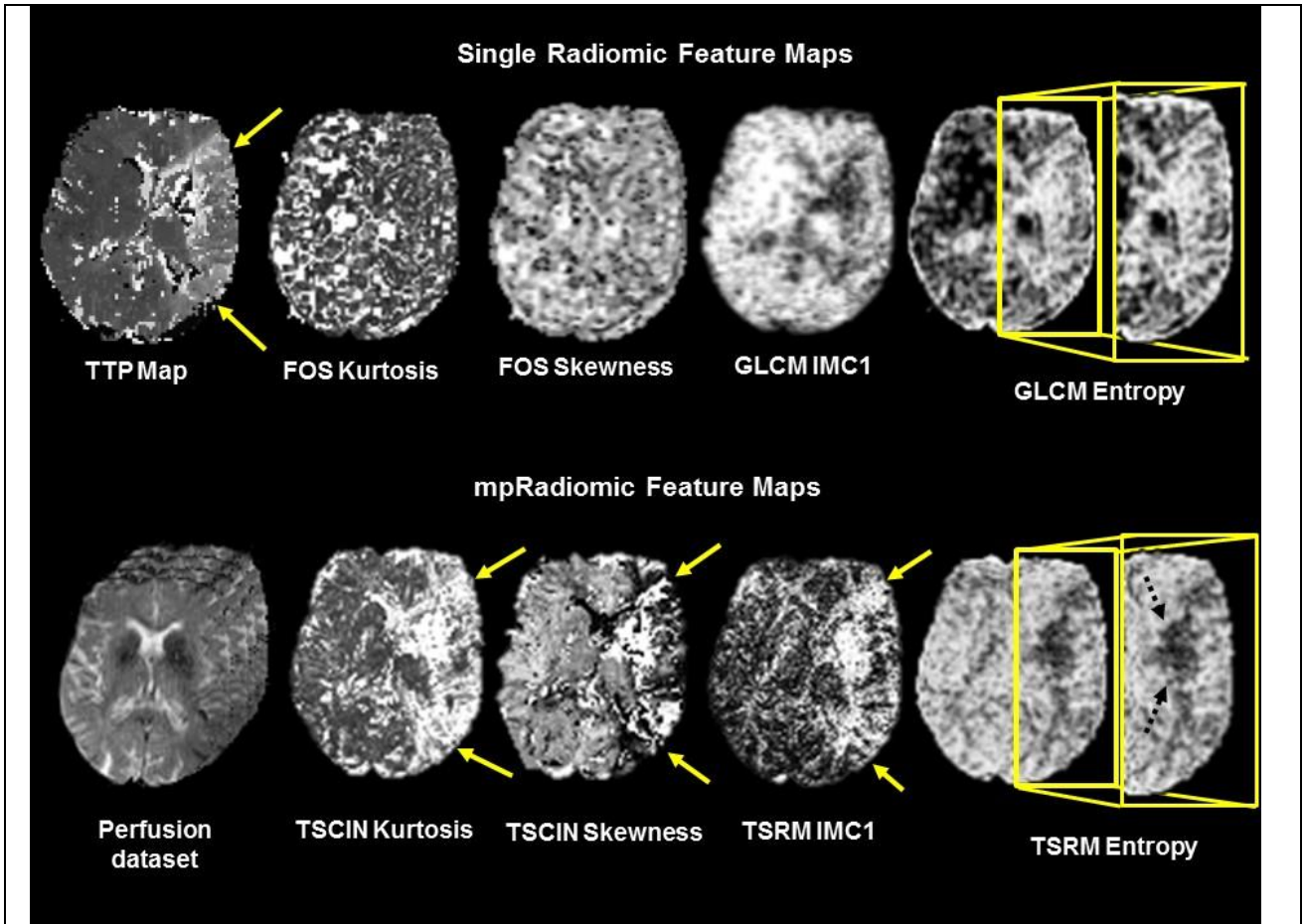

Figure 7. Illustration of radiomic feature maps (RFM) obtained from single and multiparametric radiomic analysis of an acute stroke patient with mpMRI perfusion weighted imaging. **Top Row.** Time to Peak (TTP) map from the perfusion MRI with the yellow arrow showing potential "tissue at risk". The first order (FOS) RFMs illustrate the different gray level single radiomic and co-occurrence matrix (GLCM) maps from the TTP. **Bottom Row.** MPRAD images from perfusion MRI illustrates the power mpRadiomics and the striking difference in the "tissue at risk" delineation in both the tissue signature matrix (TSCIN) and tissue signature relationship matrix (TSRM) radiomic features. The black dotted arrows show the infarcted tissue in the caudate putamen and internal capsule.

**Tables**

**Table 1**. Single and multiparametric entropy values corresponding to benign and malignant breast tumors.

| Single Parameter Entropy | Benign Tumor | Malignant Tumor | p Value | AUC |
|---|---|---|---|---|
| Entropy T1 | 4.14±0.11 | 4.66±0.06 | 0.00008 | 0.72 |
| Entropy T2 | 4.98±0.12 | 5.42±0.06 | 0.002 | 0.68 |
| Entropy b0 | 4.44±0.17 | 5.06±0.09 | 0.002 | 0.67 |
| Entropy b600 | 3.00±0.20 | 3.77±0.09 | 0.0009 | 0.67 |
| Entropy ADC | 4.90±0.12 | 5.40±0.06 | 0.0004 | 0.70 |
| Entropy Post-Contrast DCE (High Spatial Resolution) | 5.00±0.10 | 5.54±0.05 | 0.00001 | 0.75 |
| Entropy PK-DCE Pre | 4.32±0.12 | 4.65±0.05 | 0.02 | 0.62 |
| Entropy PK-DCE Post (wash-in) | 4.89±0.08 | 5.30±0.05 | 0.00006 | 0.72 |
| Entropy PK-DCE Post (wash-out) | 4.90±0.09 | 5.24±0.04 | 0.00007 | 0.69 |
|  |  |  |  |  |
| Multiparametric Entropy |  |  |  |  |
| TSPM Entropy (all Parameters) | 7.06±0.27 | 8.93±0.17 | P<0.00001 | 0.82 |
| TSPM Entropy (PK-DCE) | 7.06±0.27 | 8.92±0.17 | P<0.00001 | 0.82 |
| TSPM Entropy (High Spatial Resolution DCE) | 6.74±0.19 | 8.28±0.12 | P<0.00001 | 0.82 |
| TSPM Entropy (DWI) | 6.66±0.22 | 8.20±0.15 | P<0.00001 | 0.78 |

DWI: Diffusion Weighted Imaging
ADC: Apparent Diffusion Coefficient
PK: Pharmacokinetic
DCE: Dynamic Contrast Enhancement
FOS: First Order Statistics
TSPM: Tissue Signature Probability Matrix

**Table 2**. Single and multiparametric entropy contralateral glandular tissue values from patients with benign and malignant breast tumors.

| Single Parameter Entropy | Glandular Tissue Benign Patients | Glandular Tissue Malignant Patients | p Value |
|---|---|---|---|
| Entropy T1 | 5.29±0.11 | 5.12±0.06 | 0.20 |
| Entropy T2 | 5.37±0.10 | 5.32±0.06 | 0.68 |
| Entropy b0 | 5.19±0.24 | 4.89±0.10 | 0.27 |
| Entropy b600 | 3.46±0.24 | 3.13±0.10 | 0.20 |
| Entropy ADC | 5.27±0.28 | 5.39±0.16 | 0.71 |
| Entropy Post-Contrast DCE (High Spatial Resolution) | 5.13±0.10 | 5.00±0.06 | 0.26 |
| Entropy PK-DCE Pre | 5.24±0.12 | 5.12±0.05 | 0.38 |
| Entropy PK-DCE Post (wash-in) | 5.28±0.11 | 5.18±0.05 | 0.40 |
| Entropy PK-DCE Post (wash-out) | 5.30±0.10 | 5.24±0.05 | 0.60 |
| | | | |
| Multi-sequence entropy | | | |
| TSPM Entropy (all Parameters) | 10.93±0.34 | 10.64±0.17 | 0.46 |
| TSPM Entropy (PK-DCE) | 10.92±0.34 | 10.64±0.17 | 0.47 |
| TSPM Entropy (High Spatial Resolution DCE) | 9.17±0.17 | 9.04±0.10 | 0.51 |
| TSPM Entropy (DWI) | 9.31±0.35 | 9.06±0.18 | 0.54 |

DWI: Diffusion Weighted Imaging
ADC: Apparent Diffusion Coefficient
PK: Pharmacokinetic
DCE: Dynamic Contrast Enhancement
FOS: First Order Statistics
TSPM: Tissue Signature Probability Matrix

**Table 3.** Top multiparametric radiomic features for classification of malignant from benign breast tumors.

| S. No. | MIRAD Radiomic feature | Benign Tumor | Malignant Tumor | p Value | AUC |
|--------|------------------------|--------------|-----------------|---------|-----|
| 1 | TSPM entropy (all parameters) | 7.06±0.27 | 8.93±0.17 | p<0.00001 | 0.82 |
| 2 | TSPM entropy (DCE) | 7.06±0.27 | 8.92±0.17 | p<0.00001 | 0.82 |
| 3 | TSPM entropy (HiRes) | 6.74±0.19 | 8.28±0.12 | p<0.00001 | 0.82 |
| 4 | TSPM entropy (DWI) | 6.66±0.22 | 8.20±0.15 | p<0.00001 | 0.78 |
| 5 | TSCIN DWI max maximum | 0.44±0.02 | 0.34±0.01 | p<0.00001 | 0.77 |
| 6 | TSCIN DWI standard deviation | 0.18±0.01 | 0.12±0.00 | p<0.00001 | 0.79 |
| 7 | TSCIN DWI range | 0.34±0.02 | 0.24±0.01 | p<0.00001 | 0.79 |
| 8 | TSCIN DWI median absolute deviation | 0.13±0.01 | 0.09±0.00 | p<0.00001 | 0.78 |
| 9 | TSCIN DCE kurtosis | 2.63±0.14 | 3.37±0.08 | 0.00004 | 0.76 |
| 10 | TSCIN DCE skewness | -0.69±0.07 | -1.06±0.04 | 0.00001 | 0.75 |

**Table 4.** Multiparametric stroke radiomic values from diffusion weighted imaging in infarcted and tissue at risk.

| | Radiomic Feature | Infarcted Tissue | Tissue at Risk | White matter | Gray matter | p value (Infarcted vs Tissue at Risk) |
|---|---|---|---|---|---|---|
| **Single Parameter Radiomics (ADC map)** | Mean ADC value | 0.66±0.04 | 0.88±0.05 | 0.91±0.04 | 1.14±0.06 | 0.003 |
| | GLCM Autocorrelation | 42.36±5.09 | 71.61±7.66 | 74.59±7.30 | 111.20±11.17 | 0.01 |
| | GLCM Cluster Tendency | 2.64±0.69 | 7.29±3.81 | 1.65±0.40 | 14.15±2.27 | 0.26 |
| | GLCM Contrast | 0.83±0.12 | 1.20±0.24 | 0.61±0.11 | 5.02±0.90 | 0.19 |
| | GLCM Homogeneity1 | 0.74±0.02 | 0.72±0.01 | 0.79±0.02 | 0.54±0.02 | 0.43 |
| | GLCM Variance | 42.76±5.08 | 72.20±7.69 | 74.87±7.30 | 113.69±11.10 | 0.01 |
| | GLCM Entropy | 2.10±0.20 | 2.59±0.20 | 1.73±0.15 | 3.24±0.13 | 0.10 |
| | GLCM IMC1 | -0.18±0.03 | -0.19±0.03 | -0.18±0.02 | -0.28±0.03 | 0.75 |
| **Multiparametric Imaging Radiomics (all DWI + ADC)** | TSCM Autocorrelation | 199.66±23.35 | 133.96±17.33 | 110.95±19.87 | 137.16±18.54 | 0.04 |
| | TSCM Cluster Tendency | 206.16±29.37 | 90.61±18.50 | 73.95±33.10 | 66.26±15.38 | 0.01 |
| | TSCM Contrast | 4.20±0.94 | 1.92±0.48 | 0.58±0.08 | 5.50±0.18 | 0.05 |
| | TSCM Homogeneity1 | 0.64±0.02 | 0.70±0.02 | 0.80±0.01 | 0.58±0.02 | 0.05 |
| | TSCM Variance | 201.73±23.40 | 134.90±17.41 | 111.22±19.89 | 139.88±19.08 | 0.04 |
| | TSCM Entropy | 3.94±0.17 | 3.73±0.19 | 2.70±0.13 | 4.21±0.15 | 0.43 |
| | TSCM IMC1 | -0.43±0.02 | -0.44±0.02 | -0.54±0.02 | -0.31±0.01 | 0.74 |

DWI: Diffusion Weighted Imaging
ADC: Apparent Diffusion Coefficient
GLCM: Gray-level co-occurrence matrix
TSCM: Tissue Signature Co-occurrence Matrix
IMC1: Informational Measure of Correlation 1

**Table 5.** Multiparametric stroke radiomic values from perfusion weighted imaging in infarcted and tissue at risk

|  | Radiomic Feature | Infarcted Tissue | Tissue at Risk | White matter (contralateral) | Gray matter (contralateral) | p value (Infarcted vs Tissue at Risk) |
|---|---|---|---|---|---|---|
| Single Parameter Radiomics (TTP map) | Mean TTP value | 11.19±1.90 (sec) | 10.81±1.34 (sec) | 8.23±1.06 (sec) | 7.61±1.09 (sec) | 0.88 |
|  | FOS Entropy | 2.96±0.72 | 2.83±0.33 | 1.22±0.36 | 2.12±0.35 | 0.88 |
|  | FOS Uniformity | 0.26±0.16 | 0.21±0.05 | 0.54±0.12 | 0.36±0.09 | 0.77 |
|  | FOS Kurtosis | 0.24±0.08 | 0.40±0.03 | 3.26±0.34 | 10.72±2.52 | 0.12 |
|  | FOS Skewness | 1.64±0.57 | 1.11±0.46 | 0.14±0.33 | 0.22±1.16 | 0.49 |
|  | GLCM Correlation | 0.18±0.03 | 0.31±0.04 | 0.20±0.09 | 0.21±0.06 | 0.04 |
|  | GLCM IMC1 | -0.28±0.04 | -0.17±0.03 | -0.19±0.10 | -0.26±0.05 | 0.05 |
| Multiparametric Imaging Radiomics (all PWI series) | TSPM Entropy | 6.62±0.53 | 8.41±0.33 | 6.16±0.41 | 6.46±0.15 | 0.01 |
|  | TSPM Uniformity | 0.016±0.005 | 0.004±0.001 | 0.019±0.005 | 0.012±0.002 | 0.05 |
|  | TSCIN Kurtosis | 4.70±0.58 | 4.63±0.54 | 5.93±0.79 | 5.08±0.50 | 0.94 |
|  | TSCIN Skewness | -0.47±0.25 | -0.59±0.17 | -1.11±0.28 | -0.84±0.24 | 0.71 |
|  | TSCM Correlation | 0.76±0.06 | 0.85±0.04 | 0.86±0.05 | 0.78±0.03 | 0.24 |
|  | TSCM IMC1 | -0.23±0.03 | -0.30±0.03 | -0.37±0.05 | -0.22±0.03 | 0.11 |

TTP: time-to-peak
PWI: Perfusion Weighted Imaging
FOS: First Order Statistics
GLCM: Gray-level co-occurrence matrix
TSCM: Tissue Signature Co-occurrence Matrix
IMC1: Informational Measure of Correlation 1

**Table 6.** Multiparametric stroke radiomic values from the complete dataset consisting of diffusion and perfusion weighted imaging in infarcted and tissue at risk.

| | Radiomic Feature | Infarcted Tissue | Tissue at Risk | White matter (contralateral) | Gray matter (contralateral) | p value (Infarcted vs Tissue at Risk) |
|---|---|---|---|---|---|---|
| Multiparametric Imaging Radiomics (PWI+DWI) | TSPM Entropy | 6.65±0.52 | 8.41±0.33 | 6.16±0.41 | 6.47±0.15 | 0.01 |
| | TSPM Uniformity | 0.015±0.004 | 0.004±0.001 | 0.019±0.005 | 0.012±0.002 | 0.02 |
| | TSCM Correlation | 0.88±0.03 | 0.92±0.01 | 0.94±0.01 | 0.81±0.03 | 0.26 |
| | TSCM IMC1 | -0.33±0.03 | -0.27±0.02 | -0.42±0.04 | -0.23±0.02 | 0.07 |

PWI: Perfusion Weighted Imaging
TSPM: Tissue Signature Probability Matrix
TSCM: Tissue Signature Co-occurrence Matrix
IMC1: Informational Measure of Correlation 1